\documentclass[letterpaper, 10 pt, conference]{ieeeconf}  

\IEEEoverridecommandlockouts                              

\usepackage{graphics} 
\usepackage{epsfig} 
\usepackage{mathptmx} 
\usepackage{times} 
\usepackage{amsmath} 
\usepackage{amssymb}  
\usepackage{gensymb}
\usepackage{multirow}
\usepackage{hhline}
\usepackage{algorithm}  
\usepackage{algpseudocode}  
\usepackage{cite}
\usepackage{subfigure}       
\usepackage{graphicx}
\usepackage{xcolor}
\usepackage{url}
\usepackage{multirow}
\usepackage{bbding} 
\let\labelindent\relax
\usepackage{enumitem}
\setitemize{fullwidth,itemindent=\parindent,listparindent=\parindent,itemsep=0ex,partopsep=0pt,parsep=0ex}
\setenumerate{fullwidth,itemindent=\parindent,listparindent=\parindent,itemsep=0ex,partopsep=0pt,parsep=0ex}
\usepackage{dblfloatfix}

\title{\LARGE \bf
MultiRoboLearn: An open-source Framework for Multi-robot Deep Reinforcement Learning
}

\author{Junfeng Chen$^{1}$, Fuqin Deng$^{1}$, Yuan Gao$^{1}$, Junjie Hu$^{1}$, Xiyue Guo$^{1}$, Guanqi Liang$^{2}$ and Tin Lun Lam$^{1,2,\dagger} $
\thanks{This paper is partially supported by the National Key R\&D Program of China (2020YFB1313300) and by the special projects in key fields of Guangdong Provincial Department of Education of China under Grant 2019KZDZX1025}
\thanks{$^{1}$Authors are with the Shenzhen Institute of Artificial Intelligence and Robotics for Society.}%
\thanks{$^{2}$Authors are with the Chinese University of Hong Kong, Shenzhen.}%
\thanks{$^{\dagger}$Corresponding author: Tin Lun Lam
        {\tt\small tllam@cuhk.edu.cn}
        }%
 }
 
\begin{document}
\maketitle
\thispagestyle{empty}

\begin{abstract}

It is well known that it is difficult to have a reliable and robust framework to link multi-agent deep reinforcement learning algorithms with practical multi-robot applications. To fill this gap, we propose and build an open-source framework for multi-robot systems called \textit {MultiRoboLearn\footnote {Source code is available at https://github.com/JunfengChen-robotics/MultiRoboLearn}}. This framework builds a unified setup of simulation and real-world applications. It aims to provide standard, easy-to-use simulated scenarios that can also be easily deployed to real-world multi-robot environments. Also, the framework provides researchers with a benchmark system for comparing the performance of different reinforcement learning algorithms. We demonstrate the generality, scalability, and capability of the framework with two real-world scenarios\footnote{Our video can be found with link https://github.com/JunfengChen-robotics/MultiRoboLearn/raw/main/MultiRoboLearn.mp4} using different types of multi-agent deep reinforcement learning algorithms in discrete and continuous action spaces.

\end{abstract}

\section{Introduction}

The application of reinforcement learning to single-robot systems has gained a lot of progress in the last decades. One of the notable examples is conducted by the ETH recently, where a quadruped ANYmal robot is trained to walk on different kinds of terrains based entirely on reinforcement learning~\cite{lee2020learning}. 
However, in terms of generality, efficiency, and capability in an unstructured and large complex environment, it is also important to include support of multi-robot systems in existing robot learning frameworks. 
More specifically, complex tasks such as search/rescue, group formation control, or uneven terrain exploration require robust, reliable, and dynamic collaboration among robots, and these tasks may not be solved efficiently using traditional methods~\cite{long2018towards, mataric1997reinforcement}. As a consequence, a reasonable direction to explore learning-based methods for multi-robot systems should be considered. 
A framework that acts as a bridge to link multi-agent deep reinforcement learning (MADRL) algorithms with real-world multi-robot systems could be a tool that serves the community well.

In this paper, we propose MultiRoboLearn, an open-source framework for multi-robot deep reinforcement. This framework has two key features comparing with other frameworks.
On the one hand, compared with learning-based single-robot frameworks, this framework considers important aspects needed in the multi-robot community. For example, in multi-robot systems, it is important to consider how robots collaborate to perform tasks intelligently, and how robots communicate with each other efficiently. However, single-robot frameworks normally lack these kinds of considerations when they are designed~\cite{ todorov2012mujoco, Roboschool}. On the other hand, compared to traditional multi-robot frameworks. This work extends the system to the domain of learning algorithms, which brings more advantages to the field. It is shown that using high-dimensional sensor data received by robots, schedule planning in a dynamic, unknown, unstructured environment is difficult traditionally \cite{jaradat2011reinforcement}. Previous studies have shown that learning based approaches hold great promise to address shortcomings of traditional methods. For example, in ~\cite{long2018towards}, the authors used gradieYnt-based deep reinforcement learning algorithms to reduce the complexity of multi-robot collision avoidance problems for a large number of robots. In~\cite{huttenrauch2018local}, the authors also implemented deep reinforcement learning algorithms to solve the local communication problems between robots in a multi-robot swarm system. 

\begin{figure}[t]
\centering
\includegraphics[width=0.4\textwidth,height=2.5in]{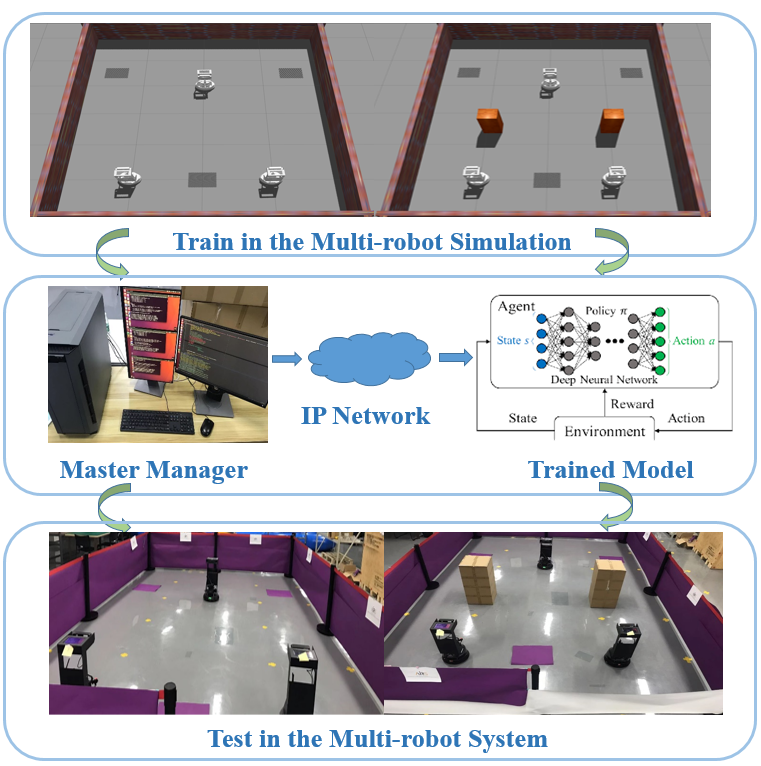}
\centering
\vspace{-2mm}
\caption{The framework for implementing multi-robot reinforcement learning algorithms. They are first learned in simulated environments and then deployed to real-world environments.
}
\label{fig:biaoti}
\vspace{-7mm}
\end{figure}

\begin{figure*}[t!]
\centering
\includegraphics[width=0.6\textwidth]{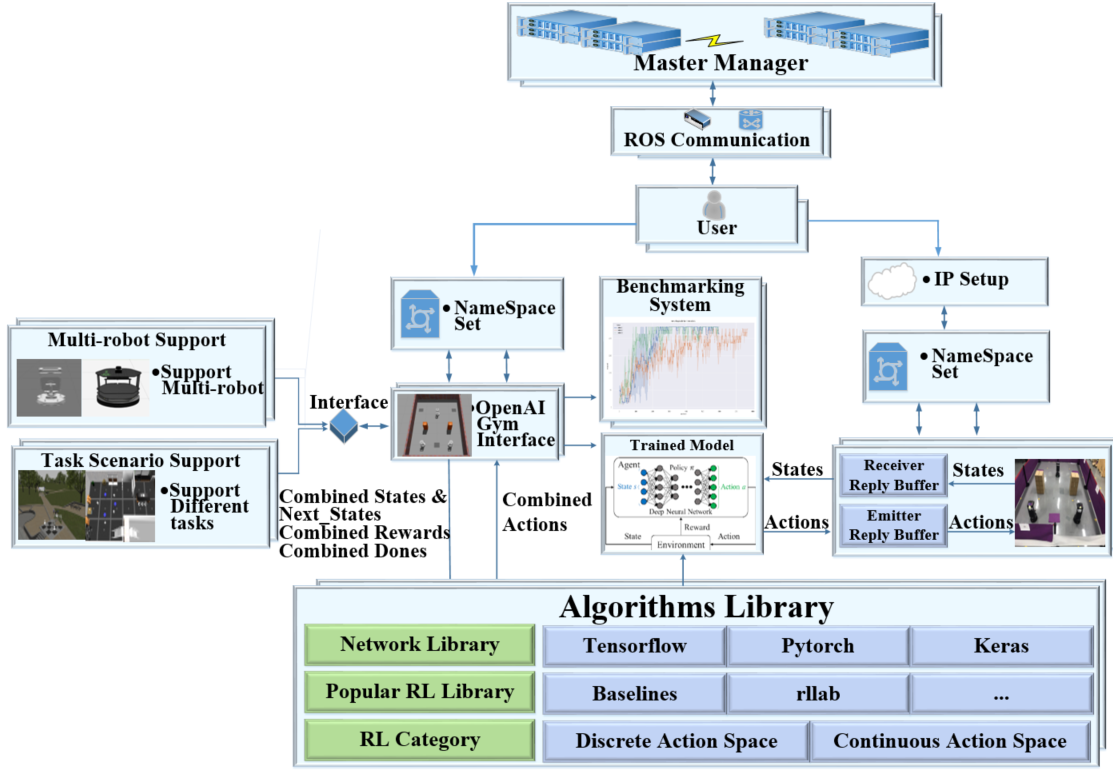}
\centering
\caption{
The details of the proposed framework structure. Its components include Master Manager, simulated environment , hardware environment and algorithm library. This framework supports learning in the simulation and test in the real-world. 
}
\label{fig:software}
\vspace{-3mm}
\end{figure*}

To the best of our knowledge, previous works were conducted without a dedicated framework to consider different aspects of implementing a learning-based multi-robot system. We hope our proposed framework serves as a foundation for research in multi-robot systems. Fig.~\ref{fig:biaoti} shows the high-level structure of our framework. During the implementation, virtual robots first need to be trained in simulated scenarios and then deployed in the real world.  Our contributions are three-fold:

\begin{itemize}
    \item  Our framework provides the community with an open-source framework for MADRL simulations and their corresponding deployment on real robots. It decouples the physical and algorithmic environments, allowing researchers to focus solely on algorithm development without knowledge of the multi-robot domain.
    \item Our framework utilizes the same code base for both simulation and real-world scenarios, providing a unified benchmark system to compare the performances of different algorithms in different situations.
    \item  Our framework also technically advances the structure of the transmitter and receiver replay buffers, enabling them to handle discretized actions and states from sensor data. This allows for consistent and smooth control of multi-robot systems with small delays. 
\end{itemize}

\section{Related Work}
\label{sec_related_work}

\subsection{Learning Based Framework for Single-robot Systems}
After the emergence of reinforcement learning in robotics, OpenAI Gym~\cite{brockman2016openai}  has become one of the standard interfaces for robot learning. 
Then It has been further combined with Mujoco~\cite{todorov2012mujoco} to provide few standardized training environments for the simulation of single robots. 
However, it does not have direct connections with real-world robotics applications. Also, the same problems have also been presented in the Roboschool~\cite{Roboschool}.
In addition, as a physical simulation engine for games, robotics and RL, Pybullet~\cite{coumans2016pybullet} provides a set of benchmarks for comparing the performance of different algorithms. However, it can only be used in simulations with a single robot. OpenAI-ros~\cite{openai_ros} inherits the interface of Gym and Gazebo, but also has the same problem as the Pybullet.
Unity Robotics~\cite{unityrobotics} provides a realistic platform where researchers can combine AI technologies to make robots complete complex tasks, however, it has not considered multi-robot version without providing any multi-robot APIs. Although Isaac Gym~\cite{makoviychuk2021isaac} has high training performance comparing with other toolkits training on CPU, it has the same issues as aforementioned toolkits which are lack of utilities linking multi-agent reinforcement learning methods to multi-robot applications.

The DeepMind Control Suite~\cite{tassa2018deepmind} provides RL environments in the simulation without physical support. It is well known that gym-gazebo2~\cite{lopez2019gym} presents an upgraded, real-world application version of gym-Gazebo, this toolkit uses a new ROS2-based software structure, but it only focuses on specific industrial robot arms, cannot be extended for other robots, and does not provide a MADRL algorithmic environment.
Recently, Deepbots\cite{kirtas2020deepbots} has designed an environment where the community can easily develop RL methods in Webots~\cite{michel1998webots}, however, they also have the above mentioned problems. Additionally, they do not consider MADRL algorithm environments or lacks the ability to be applied on real-world robots. 
PyRoboLearn~\cite{delhaisse2020pyrobolearn} integrates RL and imitation learning environments. However, it exclusively focuses on the simulation without consideration on the application of real robots.
The toolkit, robo-gym~\cite{lucchi2020robo} introduces a freely available framework allowing the use of DRL in the simulations and real-world robots. The framework is tailored for the single mobile robot and manipulator as it cannot be extend into multi-robot systems in both simulation and the real world. 

\subsection{Frameworks for Multi-agent Reinforcement Learning}
It is important to note that there is a growing body of work that focuses on MADRL from a theoretical perspective without considering how to apply these algorithms to real-world robots. Unlike the learning-based single-agent framework, to the best of our knowledge, there are no general algorithmic environments for theoretical research on MADRL or any toolkits to implement MADRL for multi-robot systems. 
Specifically, \textit{PettingZoo} \cite{terry2020pettingzoo} aims to provide a python library for multi-agent reinforcement learning akin to multi-agent version of Gym, however, this toolkit only integrates some of classic and highly-simplified environments including Atari, Butterfly, etc. 
In the game domain, Unity ML-Agents toolkit \cite{juliani2020unity} is a popular framework to design intelligent characters in the game scenarios by providing extensive, scalable, sufficient python APIs where researchers can easily train agents using reinforcement learning, imitation learning, curriculum learning, and any other methods according to the need of multiple tasks in the game. However it has the same issues as \textit{PettingZoo}. 
Although OpenAI has recently released some MADRL environments, such as the multi-agent particle environment~\cite{lowe2017multi}, hide and seek~\cite{singh-arxiv2018} and Arena~\cite{wang2019arena}, these environments cannot support the MADRL study of multi-robot systems in the real world.




\section{Proposed Framework}
\label{marl_sim}

\subsection{The Components}

\begin{figure*}[hbt!]
    \centering
    \includegraphics[width=0.8\linewidth
,height=2in]{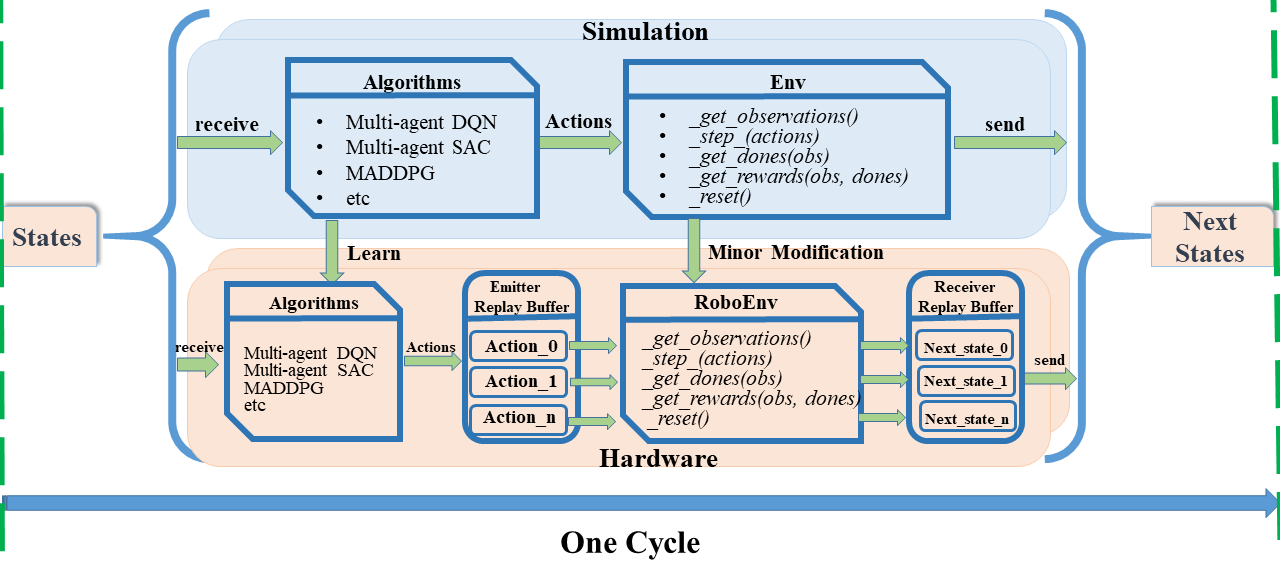}
    \caption{Process flowchart of one cycle in the simulation and hardware setup.}
    \label{fig:workflow}
    \vspace{-5mm}
\end{figure*}

We introduce our framework module by module, starting from master manager, then continuing with simulated environments, real-world environments and algorithm libraries. The relationship of these modules is illustrated in Fig.~\ref{fig:software}.

\begin{enumerate}
    \item \textit{Master Manager}: This
framework primarily uses a central control structure to drive simulated and real-world systems. The use of ROS communication structure allows for easy establishment of IP networks. This control structure facilitates users to stifle the training or testing process of simulated and real-world robots according to their experimental needs. On the other hand, this structure makes it easy to control both simulated and real-world robots with trained models.


The mManager has two functions. The first is to connect to a simulation of multi-robot systems. This makes the studying of various MADRL algorithms possible. There are two steps regarding this function. On the one hand, the task scenarios and multi-robot models are loaded through Gazebo. In this step, the mManager sets up namespaces for different robots in order to control all robots efficiently in the same simulated environment. In addition, mManager automatically sets up various services to facilitate sending and receiving of system data. On the other hand, the mManager provides an algorithmic environment by building a virtual environment using Python. Researchers can design different algorithms in this container without considering the compatibility of the Gazebo environment and the algorithmic environment. The second function is to connect and configure the physical platform. When using a multi-robot system in the real world, there are three steps to perform. First, mManager automatically establishes a local network by setting static IP addresses to add different robots and migrate them to various scenarios without having to consider network configurations. Secondly, namespace, rostopic, and rosservice are configured for each real robot autonomously. The third step in the training model, obtained from the simulation, provides a corresponding control policy for each robot to allow the real-world multi-robot systems to perform the complex tasks.


    \item \textit{Algorithm Environment}:
    This is one of the most important components in our framework, which inherits the interface of OpenAI Gym. Our implementation can be easily combined with various algorithm libraries, helping users to focus on designing the algorithm itself, without considering difficulties of connecting algorithms and environments. The main function of this component is to give some definitions and application programming interface (API). It defines collections of joint states, actions, and other necessary parameters. Also, it defines the interface to connect Gazebo to implement Markov Decision Process (MDP).

The simulated environment provides two API modules, namely multi-robot module and task scenarios module. Task scenarios module uses Gazebo as a simulator, facilitating task configuration using the model library and worlds developed by the ROS community.  For robot developers, it is easy to develop different task scenarios by designing different models and worlds. For algorithm developers who are unfamiliar with Gazebo, they also can easily combine different worlds and models to implement desired tasks. Our implemented mManager already provides a number of task types, including tasks for homogeneous and heterogeneous robots.



    \item \textit{Real-Robot Environment}: The purpose of this environment is to provide real-world multi-robot systems to test, and apply the designed algorithms. In order to achieve modularity, flexibility and simplicity, we built a real multi-robot system using real-world SPARK robot, and we designed a local network system to support their communications. Due to the loose coupling of ROS communication structure, this network can be easily connected to other robots that support ROS interface, achieving high extensibility and ease of use.
    

In this environment, an emitter replay buffer and a receiver reply buffer are implemented to discretize real world signals. This is mainly due to that in the real-world, time passes by in a continuous manner, and these buffers are needed to serve as a bridging mechanism. We designed the transmitter replay buffer to send the actions of all robots in a time interval. It is essentially a list of buffers equipped on each robot to store the commands in a queue. This list of buffers will then send the corresponding command to each robot at a fixed frequency during an operation (See Emitter Reply Buffer in Fig.~\ref{fig:workflow} for reference). 



    \item \textit{Algorithm Library}: 
To achieve modularity and generality, nManager provides a series of algorithm interfaces to support different Python-based DL libraries, such as Tensorflow, Pytorch, keras, Numpy, Pandas, etc. Also, to help MADRL researchers apply algorithms in the multi-robot system with minor modifications, we build pipelines that supports currently popular RL libraries, such as Baselines~\cite{baselines}, Spinning up~\cite{SpinningUp2018} and rllab~\cite{duan2016benchmarking}. 
For generality, the framework can support both discrete action space and continuous action space environments. 
\end{enumerate}

\subsection{The Overall Process}

This subsection describes the overall process of our framework, including simulation and real-world parts, as shown in Fig.~\ref{fig:workflow}. In the simulation, our framework first receives states of the previous step, and then subsequently sends them to the algorithms. After that, the algorithms select some actions and then send them to the corresponding robot in real world applications. After actions are executed, the next states can be obtained from the environment. In real world, the algorithm selects the actions and passes them to the emitter replay buffer in order, then passes the actions to the corresponding robots in the RoboEnv. The RoboEnv collects the state of each robot at the same time by using receiver replay buffer and packages all states together into a buffer. All the process mentioned above is called a cycle.

The difference between our framework and single-robot frameworks is that we have to consider the current state of all robots in a cycle time. For example, we have to collect the current states of all robots, which is called a joint state of multi-robot system.  According to robots' relationships (e.g. cooperative, competitive or semi-cooperative, semi-competitive), we will also have to set a joint rewards structure for all the robots. In real-world multi-robot systems, concurrency is a key to the smooth execution of the command, data collection, and collaboration. Commands will be executed in a continuous period of time, which is allowed for the single robot. However, for a multi-robot system, each robot executes the corresponding command sequentially, which leads to inconsistencies and can cause communication delays or blockages. This framework builds a list of emitter and receiver replay buffers by setting an execution time for the commands. In operation, our framework sends the stored instructions to the corresponding transmitter buffer, which then sent them to the real robot. Each robot can send and collect data at the same time, enabling the multi-robot system to complete its tasks smoothly.




\subsection{Generality and Extensibility of the Framework}

\begin{enumerate}
    \item \textit{Adding Robots in Simulation \& Real World}: In the simulation, the new robot model can be easily integrated into the simulation through URDF. The multi-robot system's network can be configured with only minor code changes by using the mManager.
Regarding the real world scenario, robots can be added into multi-robot systems by creating IP in the local network and creating namespaces for new robots. Also, each robot requires a specific transmitter and receiver replay buffer. However, these hardware settings can be added by making a small modification to the original code.

    \item \textit{Using Algorithms}: New
Reinforcement learning algorithms, including single-agent and multi-agent ones can be easily developed and utilized due to our framework's integration with most of the deep reinforcement learning libraries. For MADRL, our framework supports various training methods, such as independent network training, centralized training and  decentralized execution, etc.

\end{enumerate}




\begin{figure}[t!]
\centering
\subfigure[The first task is three SPARKs navigating goal areas without obstacle.]{
\centering
\includegraphics[width=0.46\linewidth,height=1.6in]{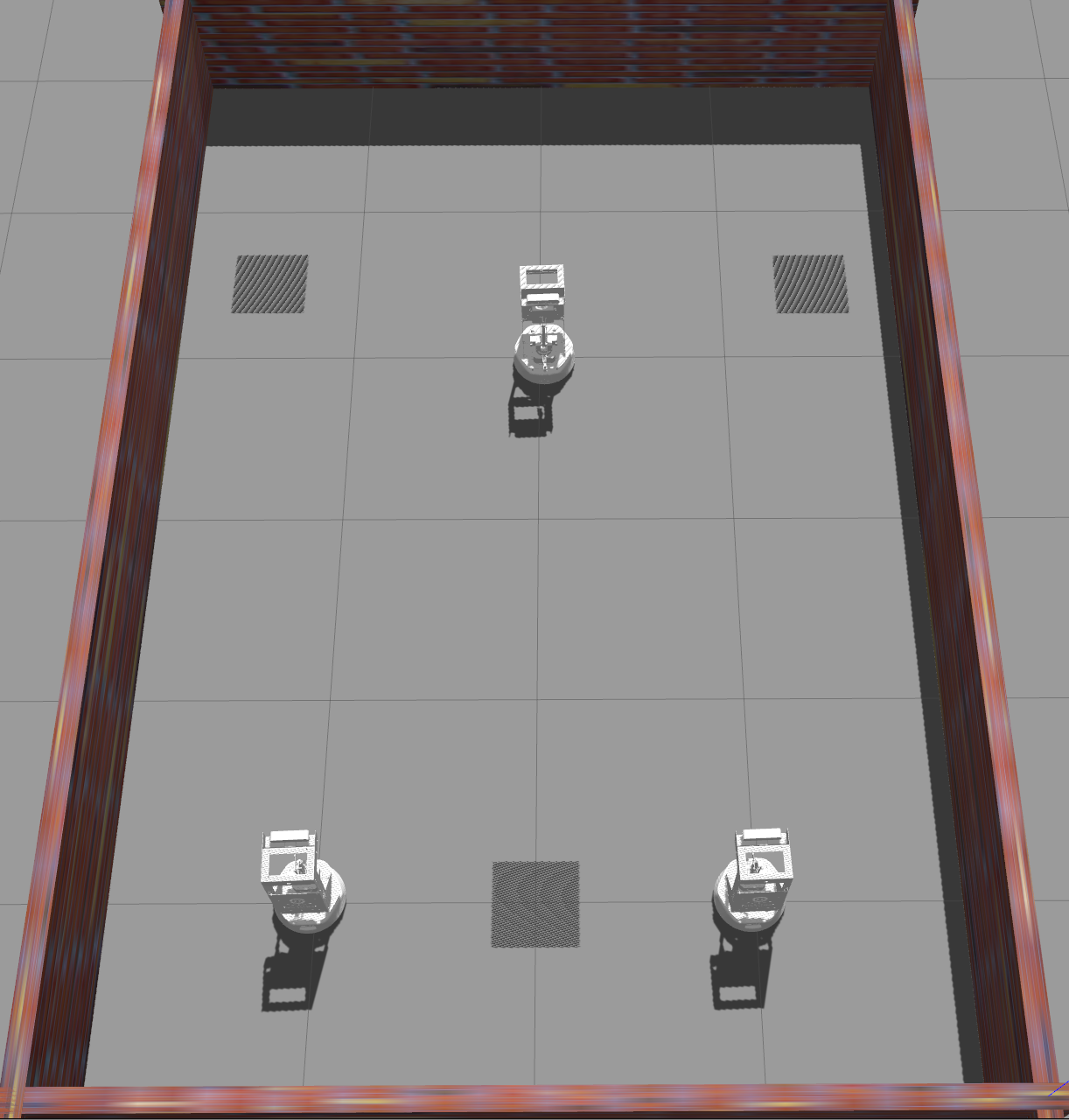}
}
\subfigure[The second task is three SPARKs navigating goal areas with obstacle.]{
\centering
\includegraphics[width=0.46\linewidth, height=1.57in]{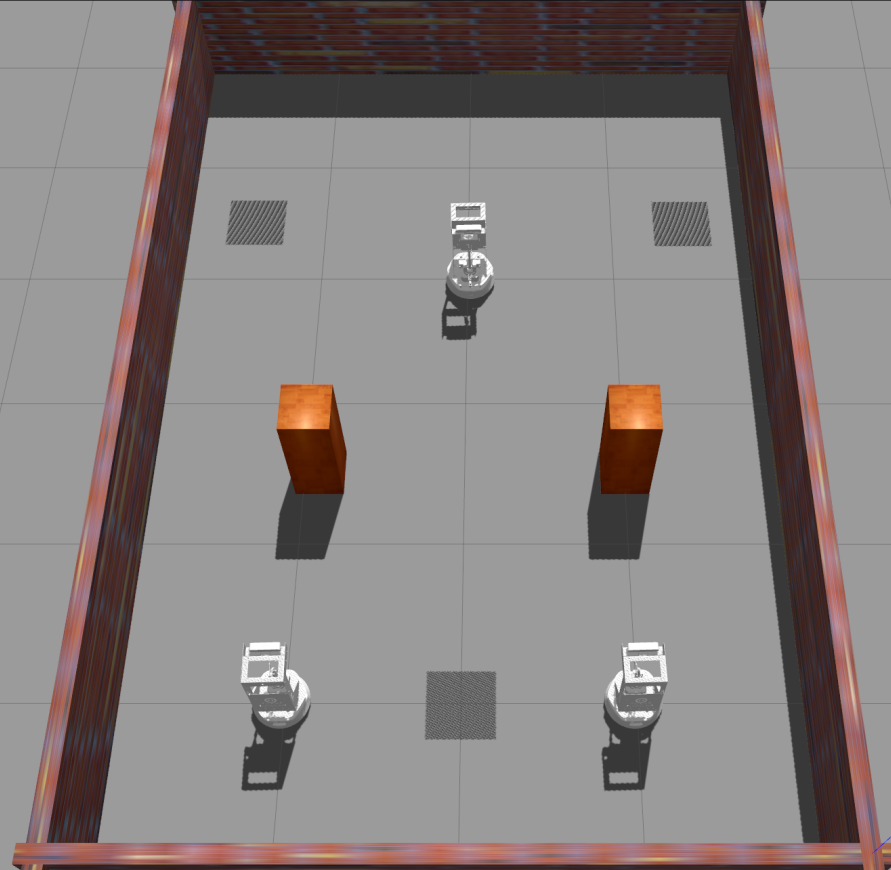}
}
\centering
\vspace{-2mm}
\caption{
The figures show simulated environments for training multi-robot reinforcement learning algorithms.
(a) shows three SPARKs with white color and three grey squares represents goal areas. (b) has the same configurations as (a) with two orange rectangular boxes as obstacles.  
}
\label{fig:experiment}
\vspace{-4mm}
\end{figure}

\section{APPLICATION}
\label{applications}


To illustrate the functionality of our framework for multi-robot learning tasks, we show our framework's performance in two different scenarios, each with two different cases. One of them is the discrete action space case and the other one is the continuous action space case. 

In the first scenario, we consider a task where three SPARKs{\footnote{https://github.com/NXROBO/spark}} collaborate to earch and chase a Laikago robot{\footnote{https://www.unitree.com/}}  while avoiding obstacles and each other. The SPARK is a differentially driven mobile robot that is currently widely used in research and ROS education. It is well supported by ROS at both hardware and software levels, and the multiple extensions provided by the ROS community make it a suitable test-bed for our framework. 

The second scenario also has three SPARKs and its main difference comparing with the first scenario is the presence of two boxes as static obstacles. In both scenarios, we use homogeneous robots with same configurations. With our framework, it is easy to integrate different configurations of the robots due to accessibility and modularity based on ROS communications. We used the same algorithms to validate our framework.


\begin{figure}[hbt]
\centering
\subfigure[Three SPARKs reaching goal areas in an obstacle-free environment.]{
\centering
\includegraphics[width=0.47\linewidth]{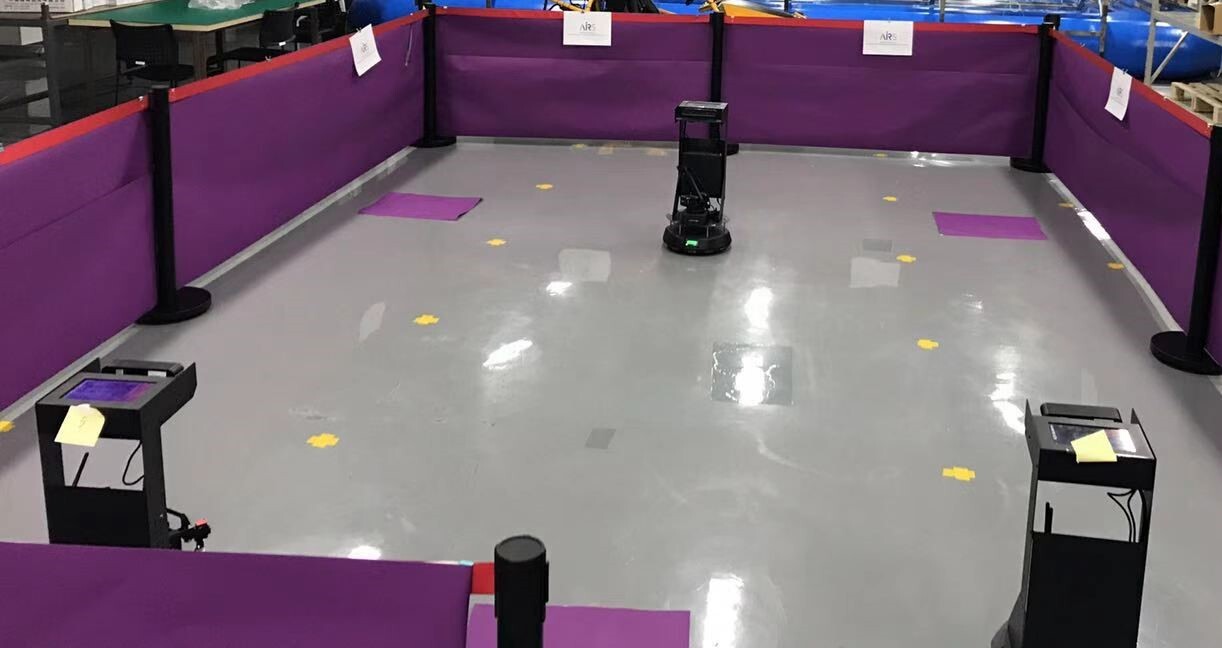}
}
\subfigure[Three SPARKs reaching goal areas in an environment with two obstacles]{
\centering
\includegraphics[width=0.42\linewidth]{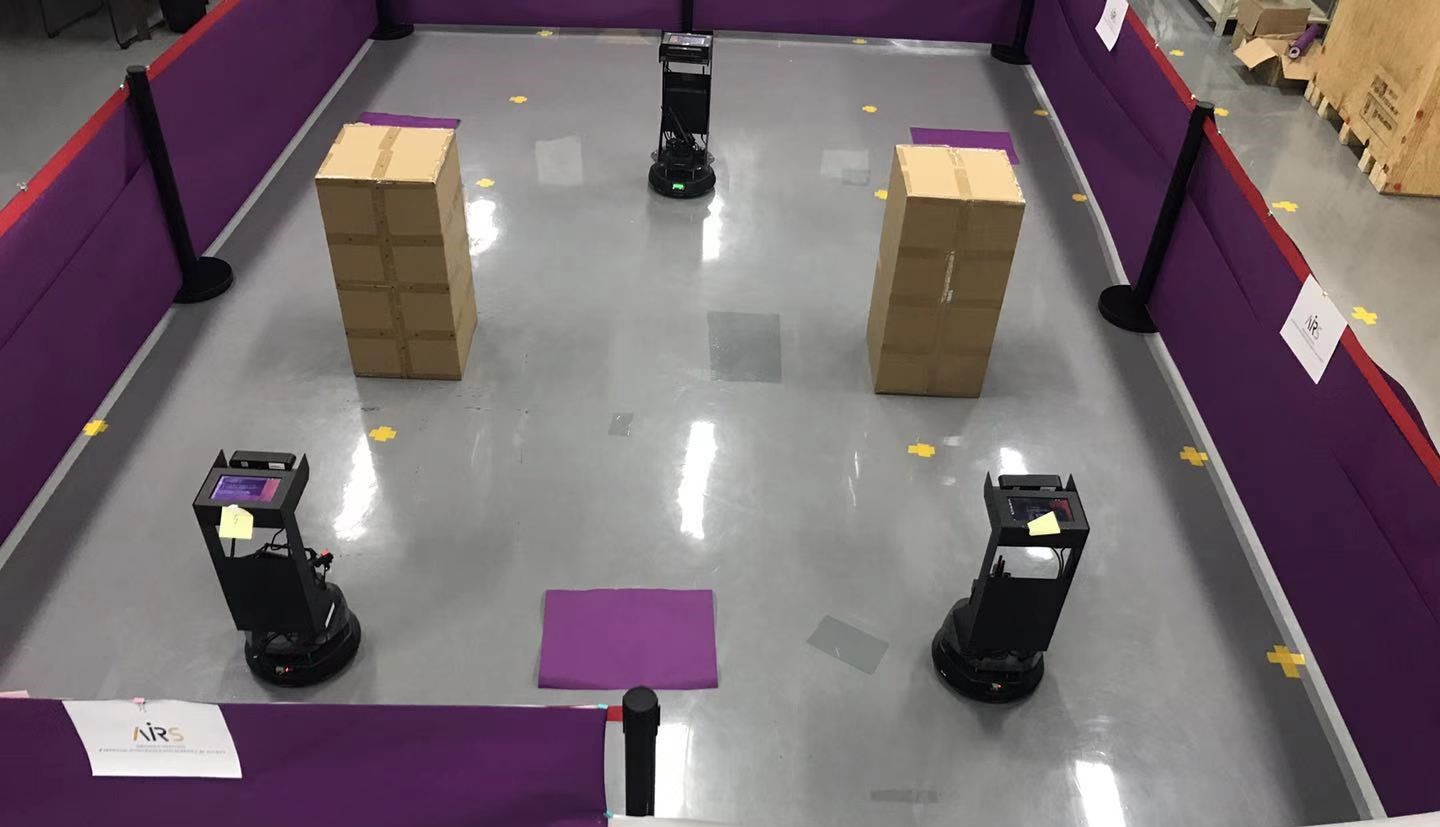}
}
\centering
\vspace{-2mm}
\caption{
Real-world environments for deploying multi-robot reinforcement learning algorithms in our experiments. They have the same configuration and scale size with the virtual environments.
}
\label{fig:experiment scenario}
\vspace{-4mm}
\end{figure}

Fig.~\ref{fig:experiment} illustrates our scenarios in the simulated environments. In the figure, white objects represent the simulated SPARK robots, red blocks represent the obstacles and grey areas represent the goals that robots need to move to. The first scenario is illustrated in Fig.~\ref{fig:experiment} (a) on the left side and the second scenario is illustrated in Fig.~\ref{fig:experiment} (b) on the right side. The corresponding real-world scenarios are shown in Fig.~\ref{fig:experiment scenario}. In the figure, black objects represent real-world SPAKR robots, orange blocks represent obstacles, and purple areas are the goal areas.

\subsection{Problem Description}
\label{problem_description}
\begin{enumerate}
    \item \textit{Mobile Navigation of SPARKs in obstacle-free environment}: In this scenario, the goal is that three SPARKs collaborate to reach each goal position while avoiding collisions during navigation, as shown in Fig.~\ref{fig:experiment} (a).
All three SPARKs have the same configuration in the dynamic model and the same parameter settings in the hardware. On the one hand, in order to detect static and dynamic obstacles, SPARK is equipped with a laser sensor that scans 360 degrees in a two-dimensional plane. In addition, to determine the current position of each robot, SPARK is equipped with an odometer that provides the local coordinates of each robot with an average error of 0.03 m over a distance of 1 m. The whole scene is placed in a room of $4\times6$ square meters. 
The joint observations are designed as a list container that collects each robot's observation with 17 values. The first two values from odometers are the robot's position in the global coordinates of the scenario. Note that all the robots are positioned in the same global coordinates over episodes. The remaining 15 values are the measured distances of the laser scanner evenly placed around each robot. The sparse processing of the laser scans may lead to the inability to perceive other dynamic robots. We deal with this problem in two ways. The first way is that in the simulation, we enlarge the collision model volume of each robot's laser scanner. The other way is to preprocess the received data from the laser scanner of real robots. To ensure the consistency between simulation and real-world laser data, we add some Gaussian noise to the laser scanner in the simulation. Also, to reduce the complexity of learning, these data from both simulation and hardware are downsampled.

In the case of discrete action space, the action is a series of discrete numbers, essentially the speed controller of the mobile robot. In the case of continuous action space, the action consists of two continuous values, namely linear velocity and angular velocity.


The difference between RL and MADRL in our case is the reward function. It should be noted that MADRL needs to consider agents' relationships, such as cooperation, competition, and half-cooperation-half-competition. In these scenarios, we consider a scenario where robots need to collaborate to perform tasks. The base reward received by the robot at each step is propositional to the distance the robot moved during last step. For all robots that reaches their goal areas simultaneously, they will all receive a large positive reward. If the robot reaches its goal area by itself, it will receive a small positive reward. Moreover, in the case of the collision, they will receive a small negative reward. 

    \item \textit{Mobile Navigation of SPARKs in obstacle
environment}: 
In this scenario task, the difference, comparing with the previous scenario, is that there are some static obstacles.
In robotics, navigation and obstacle avoidance are common research topics in the community, so the purpose of this task scenario is to demonstrate that this framework is applicable to a wide range of task scenarios and can solve common problems in the field of robotics.
\end{enumerate}

\begin{figure*}[t!]
\centering
\subfigure[SAC in obstacle-free environment]{
\centering
\includegraphics[width=0.45\linewidth]{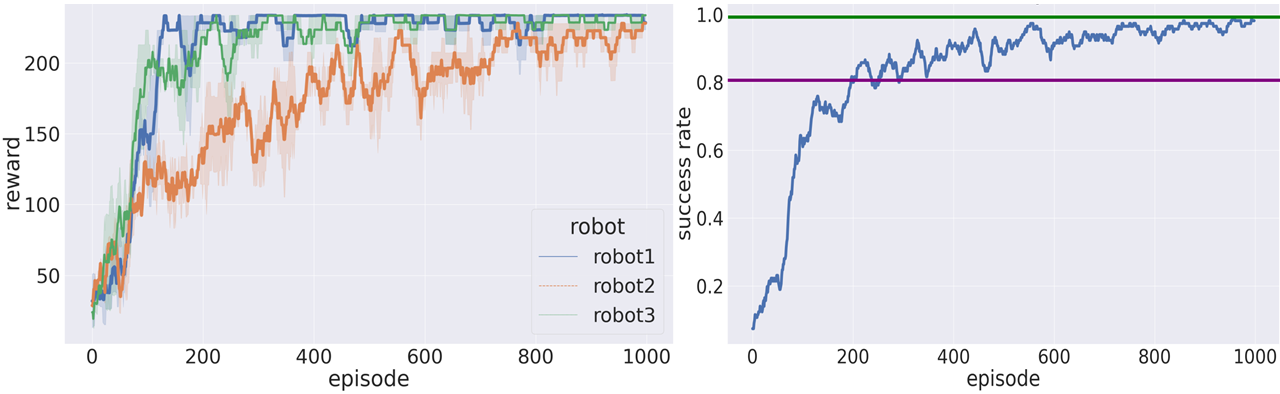}
}
\subfigure[SAC in obstacle environment]{
\centering
\includegraphics[width=0.45\linewidth]{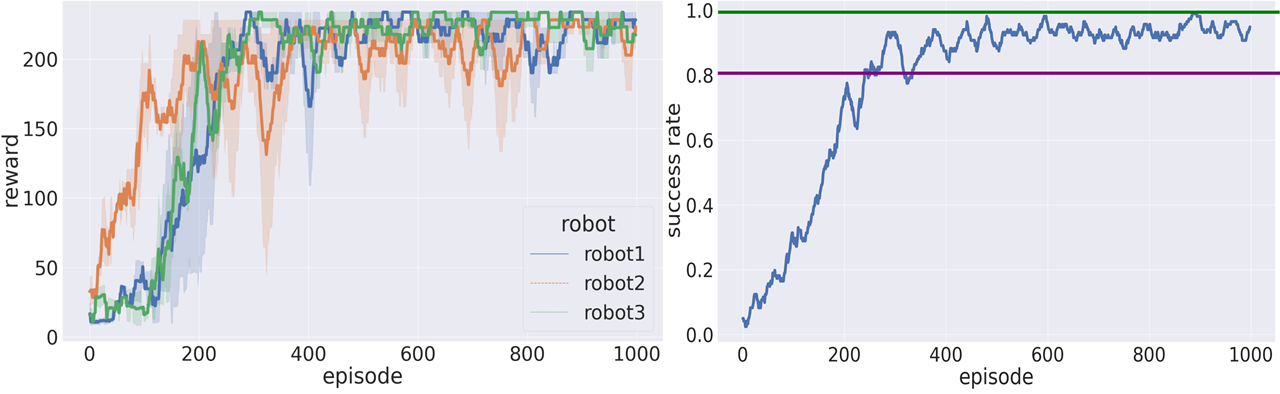}
}
\subfigure[DQN in obstacle-free environment]{
\centering
\includegraphics[width=0.45\linewidth]{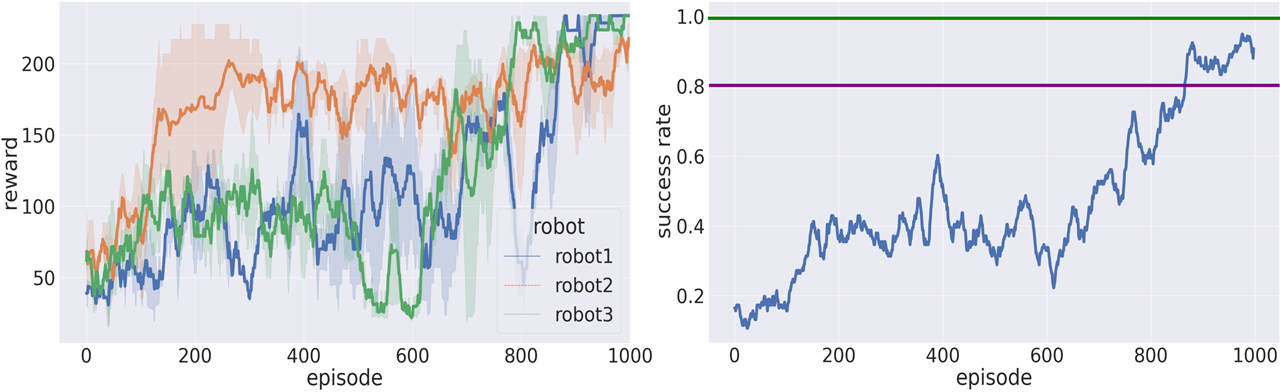}
}
\subfigure[DQN in obstacle environment]{
\centering
\includegraphics[width=0.45\linewidth]{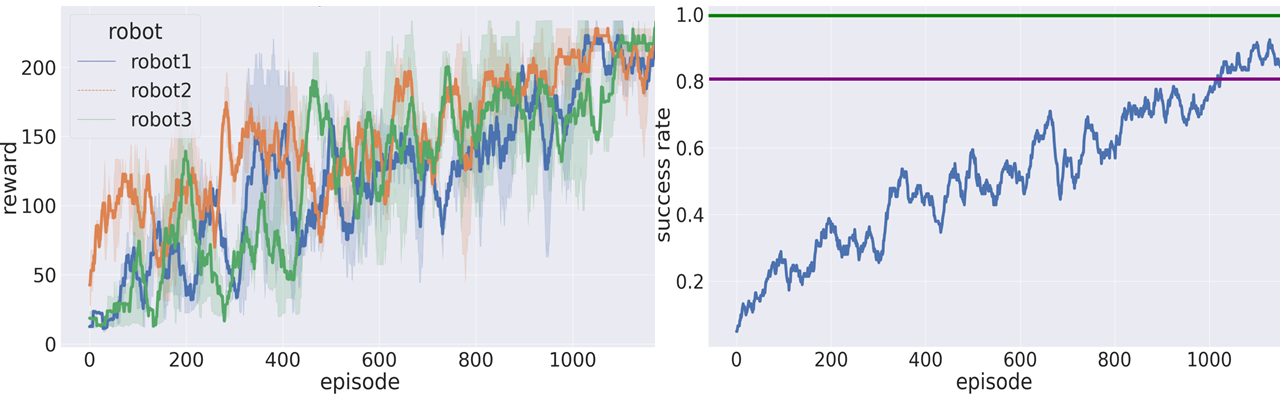}
}
\centering
\vspace{-2mm}
\caption{Cumulated reward and success rate graphs are obtained in above environment using SAC and multi-agent DQN. (a) shows convergence result of SAC in the obstacle-free environment, including reward-episode and success rate graphs. (b)shows convergence result of SAC in obstacle environment, including reward-episode and success rate graphs. (c)shows convergence result of multi-DQN in the obstacle-free environment, including reward-episode and success rate graphs. (d)shows convergence result of multi-DQN in obstacle environment, including reward-episode and success rate graphs.}
\label{fig:experiment_data}
\vspace{-3mm}
\end{figure*}

\subsection{Reinforcement Learning Algorithms}
\label{multi_agent_reinforcement_learning}

In this subsection, we provide a series of cases showing a proof of concept by using two kinds of algorithms in two cases to demonstrate the capability of this framework. In the case of continuous action space, SAC is chosen by improving SAC algorithm \cite{pytorch_sac}. In the case of discrete action space, multi-DQN same as \cite{stable-baselines} is well employed.

The SAC algorithm, contains prioritized experience replay, double DQN network to improve the original algorithm's performance. Hyperparameters are chosen according to the benchmark for \cite{pytorch_sac} with some minor modifications. 
The multi-DQN algorithm adopts Dueling-DQN. This algorithm includes the proposed prioritized experience replay, and sum tree \cite{stable-baselines} to improve performance. Hyperparameters are mainly derived from \cite{stable-baselines} with minor modification.

\subsection{The Hardware Setup}
\label{the_setups}

\begin{enumerate}
    \item \textit{Computer Setup for Training and Test}: As seen in Fig.~\ref{fig:biaoti}, mManager (essentially a computer with 20 CPU cores)  provides the functionality of switching between specific modes. When switching to the simulation mode, mManager can be utilized in the learning process. When switching to the real-world mode, mManager serves to provide control policies for all robots during test process.
    \item \textit{Real-world Setup}: In
the real-world experiments, we construct two task scenarios in the laboratory, which resembles the simulated scenarios to reduce the gap between simulation and the real world, improving the deployment accuracy of trained model in the real-world robots. In Fig.~\ref{fig:experiment scenario}, standing barriers are utilized as walls and restricted areas. In addition, two boxes are used as obstacles. The whole scene maps to $4\times6$ square meters. The trained model, receiver and emitter replay buffer are running on the mManager and connected to SPARKs' local network using Wi-Fi. To ensure consistency of all robots, commands are sent in 20 ms and executed in 100 ms.
\end{enumerate}

\subsection{Experiment Results}
\label{experiment_results}
In this subsection, our framework is evaluated using two types of algorithms in two scenarios. During the experiment, three agents are first trained in the same simulated environment. When the training process is finished, the trained model, namely control policies, is tested in both simulated and real-world environments. We show that the trained model can be directly deployed in real-world scenarios without any modification or any further training in real-world robots. See the accompanying video for the demonstration of the two scenarios in both simulation and the real world.

\begin{enumerate}
    \item \textit{Results for SAC in two Scenarios}: 
Three agents are all trained in an obstacle-free environment and obstacle environment by using a SAC algorithm with the same hyperparameters. In the obstacle-free scenario, three agents are trained to perform the task of navigation after 2000 episodes in the simulation. However, after 600 episodes, reward curves starts to converge (see Fig.~\ref{fig:experiment_data} (a) for reference). The convergence criterion for multi-robot systems is different compared to the single-robot case. In our experiments, for a single robot, the model is considered to have converged when the success rate exceeds 80\% for 20 consecutive episodes. However, a multi-robot system is considered converged only when all the robots in the multi-robot system converge. In an obstacle-free scenario, the success rate remained between 80\% and 100\% after more than 600 episodes of training. In the other scenario with obstacles, we noticed that it was more difficult to reach convergence than in the obstacle-free scenario. In fact, it is after 800 episodes that the reward curve starts to converge and the final success rate stayed between 80\% and 100\%.

To have a final evaluation, the trained models are tested in both simulated for 200 episodes as well as real-world environments for 50 episodes. In the obstacle-free scenario, the trained model had 91.5\% success rate in the simulation and had 82.0\% success rate in the real world. In the second scenario, the trained model had 92.0\% success rate in the simulation and had 75.0\% success rate. Although the success rate of a real-world experiment is lower than that of a simulation experiment, it is well demonstrated that this framework can be directly applied in a multi-robot system.


    \item \textit{Results for Multi-DQN in two Scenarios}: The multi-DQN algorithm is also trained in the two task scenarios with the same task-related hyperparameters. All the agents are trained over 2000 episodes in the simulation. In the obstacle-free scenario, the training converged after 900 episodes with a success rate between 80 \% and 100\% (see Fig. \ref{fig:experiment_data} (c)). In the other scenario with obstacles, after training over 1000 episodes, all the agents can reach goal areas with a success rate between 80\% and 100\% (see Fig. \ref{fig:experiment_data} (d)). A final evaluation of 200 episodes is also performed in simulation and the evaluation of 50 episodes is performed in the real-world scenarios. In the obstacle-free environment, the trained model reached 82\% success rate in the simulation and had 52.0\% success rate in the real world. In the second environment, the trained model had 76.5\% success rate in the simulation and had 40.0\% success rate.

\end{enumerate}

\section{CONCLUSION AND FUTURE WORK}
\label{conclusions and future work}
In this work, we proposed and built an open-source framework for multi-robot deep reinforcement learning named MultiRoboLearn. 
It provides a standardized way to apply MADRL on Gazebo. The results show that our framework can train/test multi-robot agents in simulation, benchmark algorithms continuously, and deploy the trained algorithms to robots in the real-world.Our goal is to have a growing toolkit that will serve as a solid foundation for developing research in the field of MADRL for multi-robot systems.


\bibliographystyle{IEEEtran}   
\bibliography{reference.bib}

\end{document}